\documentclass[runningheads]{llncs}
\usepackage[T1]{fontenc}
\usepackage{graphicx,verbatim}
\usepackage[square,numbers,sort&compress,sectionbib]{natbib}
\usepackage{amsmath}
\usepackage{amssymb}
\usepackage{booktabs}
\usepackage{float}
\usepackage{hyperref}
\usepackage{siunitx}
\usepackage{xcolor}
\usepackage{enumitem}
\usepackage{xspace}

\newlist{inlist}{enumerate*}{1}
\setlist[inlist]{label=\xspace(\arabic*)}
\newlist{inlistalpha}{enumerate*}{1}
\setlist[inlistalpha]{label=\xspace(\alph*)}

\makeatletter
\DeclareRobustCommand\onedot{\futurelet\@let@token\@onedot}
\def\@onedot{\ifx\@let@token.\else.\null\fi\xspace}
\def\eg{\emph{e.g}\onedot}

\def\etal{\emph{et al}\onedot}

\begin{document}
%
\title{Deformable CT-US Registration via Anatomy-Aware Implicit Neural Representations}


\titlerunning{Deformable CT-US Registration}
%

\author{
Agnieszka Lach\inst{1,2} \and
Magdalena Wysocki\inst{2} \and
Feng Li\inst{2} \and
Mohammad Farid Azampour\inst{2} \and
Benjamin D. Killeen\inst{2} \and
Felix Ginzinger\inst{1} \and
Mathias Braun\inst{1} \and
Philipp Steininger\inst{1} \and
Heinz Deutschmann\inst{1} \and
Nassir Navab\inst{2}
}

\authorrunning{A. Lach et al.}

\institute{
medPhoton GmbH, Salzburg, Austria \and
Computer Aided Medical Procedures, Technical University of Munich, Munich, Germany\\
\email{agnieszka.lach@tum.de}
}

\maketitle              
\begin{abstract}
  Slice-to-volume registration between ultrasound (US) and preoperative computed tomography (CT) imaging would enhance many minimally invasive interventions, for example by locating soft tissue structures intra-operatively that are discernible in CT. While optical tracking enables initial rigid registration, contact from the probe induces soft tissue deformations that inhibit accurate alignment. In this work, we introduce a deformable CT-ultrasound registration framework that incorporates anatomical priors derived from CT to improve registration under deformation. Rigid registration is first established using a robot-assisted optical tracking system, after which a deformable transformation is estimated using a sinusoidal implicit neural representation (SIREN) optimized per frame. Tissue stiffness is approximated from CT-based HU values and used as spatially varying regularization, suppressing deformation in rigid structures such as bone while allowing more flexibility in soft tissue.
  Two additional constraints capture the physics of probe contact: a contact-zone displacement prior that drives the displacement field to compress tissue below the probe face, and a fan-geometry regularization term based on beam direction and convex transducer field of view. Model parameters are optimized with a normalized gradient field (NGF). The proposed approach improves alignment over rigid initialisation by 17\% and outperforms classical deformable baselines while maintaining near-zero topological folding.



\end{abstract}
\section{Introduction}\label{sec:introduction}

\begin{figure}[t]
  \centering
  \includegraphics[width=\linewidth]{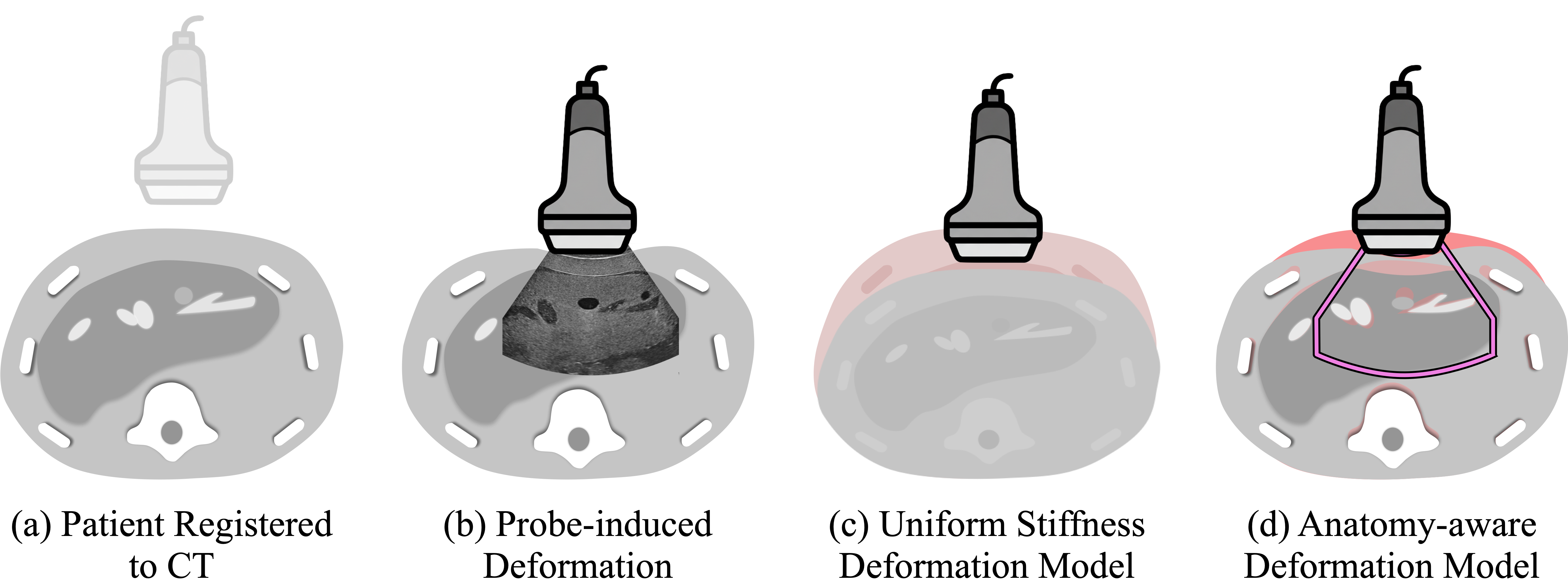}
  \caption{To enable precise anatomical alignment between tracked ultrasound and CT (a), it is necessary to account for probe-induced soft-tissue deformation (b). Existing methods are failing to incorporate anatomical constraints, allowing for non-physically plausible deformations (c) (exaggerated for visual clarity). We propose a deformable registration framework that incorporates known physical constraints and anatomical priors to constrain the deformation field, improving alignment while maintaining physical plausibility (d).}
  \label{fig:overview}
\end{figure}

Tracked intra-operative ultrasound (US) provides real-time, quantitative information about soft tissue dynamics, enabling high precision guidance for interventional procedures~\cite{pavone2024ultrasound}.
To overcome US-specific limitations such as acoustic shadowing and shallow penetration, there is growing interest in combining US with preoperative and intra-operative computed tomography (CT) imaging~\cite{li2025robotic, masoumi2021multimodal, xu2014percutaneous, li2026cbctalive}, thereby accessing 3D anatomical context dynamically. However, even with accurate rigid co-registration, \eg, via optical tracking, spatial misalignment persists due to probe-induced soft-tissue deformation during US acquisition. Strong contact between the probe and soft tissue is necessary to achieve sufficient image quality, but the resulting pressure introduces spatial inconsistencies between the US and CT volumes, as shown in Fig.~\ref{fig:overview}a-b. This prevents precise anatomical alignment for high precision interventions such as liver tumor ablation or biopsy~\cite{xu2014percutaneous}. While electromagnetic tracking can estimate deformation directly~\cite{xu2014percutaneous}, it introduces significant complexity to clinical workflows and may be susceptible to interference from metallic objects. Thus, there is a need for deformable registration methods that reconcile tracked US with CT volumes while accounting for probe-induced tissue deformation.


Existing methods lack anatomy-aware constraints and thus risk non-physically plausible deformations (Fig.~\ref{fig:overview}c): physical priors are either impractical—e.g.\ Geng~\etal~\cite{geng2025padreg} require contact-force sensors unavailable in CT-US—or, in recent learning-based work, only folded into the training objective of convolutional~\cite{azampour2023anatomy, jian2022weakly} or implicit neural representation (INR)~\cite{wolterink2022implicit,Sideri-Lampretsa2024SINR} models. INRs offer a compelling per-pair alternative to amortized learning, motivating our anatomy-constrained INR formulation.


Here, we propose a CT-US registration framework that incorporates known physical constraints and anatomical priors into an INR of deformation.
Our primary contribution is the method which uses three anatomy-aware loss terms tailored to the CT-US slice to volume probe-contact setting, which are integrated with a NGF-based multi-modal similarity metric and Jacobian regularization~\cite{wolterink2022implicit}.
First, we encode tissue stiffness as a spatially varying regularization term derived from CT-based segmentation without requiring any finite-element solver. Second, we introduce a contact-zone displacement prior that enforces the physics of probe-induced surface compression by penalizing residual tissue in the crescent region between the probe face and the US fan arc. Finally, we tailor the optimization strategy to the fan-shaped, partial field of view of a convex transducer through beam-aligned directional regularization and an angular-tapered boundary weight that attenuates displacement toward the lateral fan horns where beam quality is weakest. We evaluate our approach on two datasets of robot-assisted US-CT acquisitions of abdominal phantoms, demonstrating superior alignment compared to both rigid initialization (up to 17\% lower TRE) and classical deformable baselines.

\section{Methods}
\label{sec:methods}

\begin{figure}[t]
  \centering
  \includegraphics[width=\textwidth]{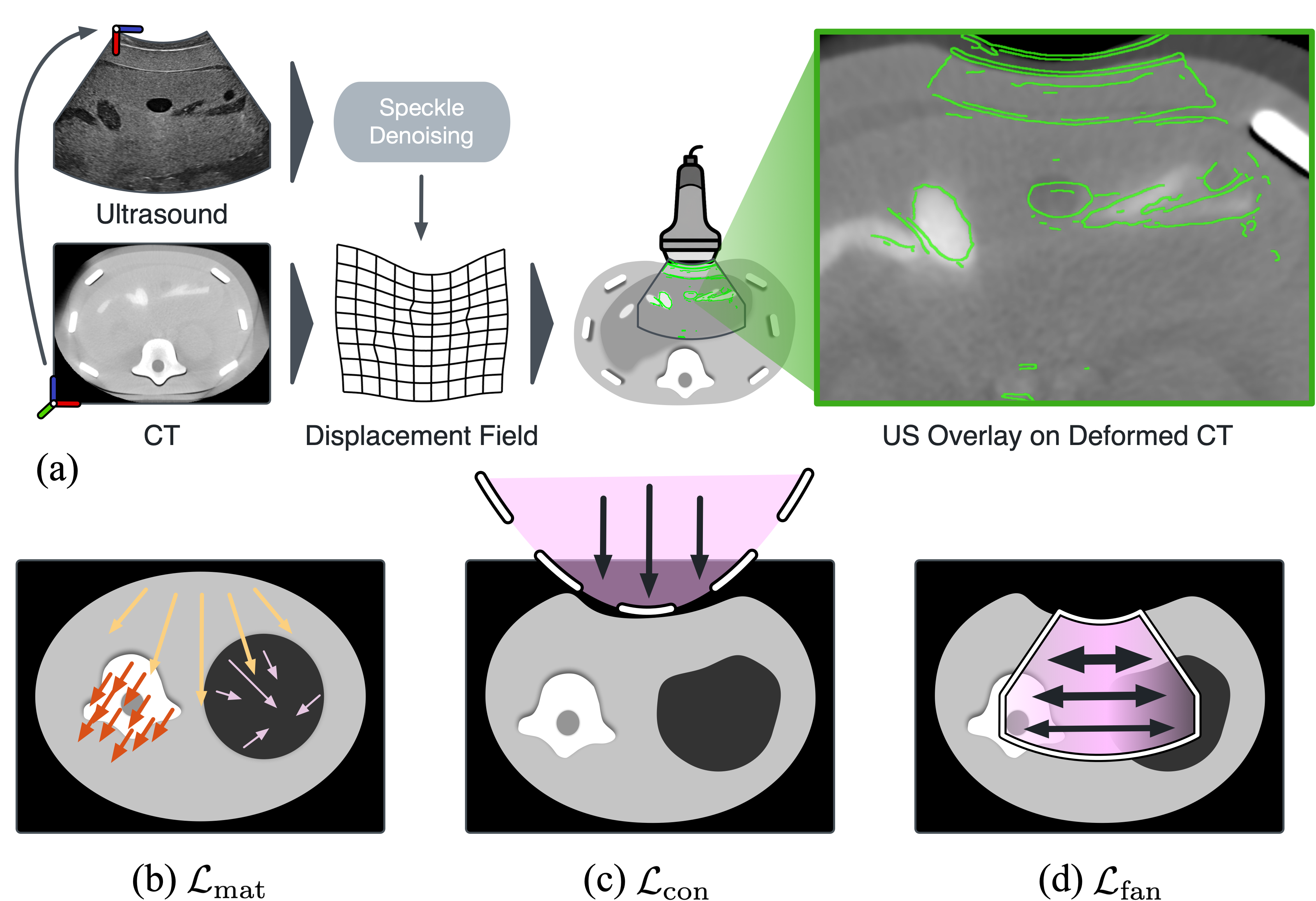}
  \caption{Proposed non-rigid CT volume  to US slice registration pipeline. The workflow transitions from initial rigid alignment to a displacement field estimation. The deformation is governed by three novel regularization terms: (a) $\mathcal{L}_{mat}$ for material stiffness preservation, (b) $\mathcal{L}_{con}$ for contact-zone interface modeling, and (c) $\mathcal{L}_{fan}$ for FOV-constrained spatial smoothing.}
  \label{fig:method}
\end{figure}


Given a 2D US frame $I_{\mathrm{US}}\!\in\!\mathbb{R}^{H\times W}$ acquired by a tracked convex transducer, we seek a dense displacement field $\mathbf{d}\colon\mathbb{R}^3\!\to\!\mathbb{R}^3$ that maximizes the alignment of anatomical structures between the 3D CT volume $I_{\mathrm{CT}}$ and the US frame for all pixels $\mathbf{x}$ inside the US fan $\Omega_{\mathrm{fan}}$. Here, $I_{\mathrm{CT}}$ is resampled to the US pixel grid via optical-tracking calibration~\cite{li2025robotic}. We parameterize $\mathbf{d}$ as a sinusoidal implicit neural representation~(SIREN~\cite{sitzmann2020implicit}) optimized \emph{per frame}. Similarity between the two modalities is measured with a normalized gradient fields (NGF) metric, and the Jacobian of $\mathbf{d}$ is regularized following Wolterink~et~al.~\cite{wolterink2022implicit}. As out-of-plane motion is unobservable from a single US slice, the displacement is constrained to the US imaging plane, yielding a 2.5-D warp.

\subsection{CT-Derived Stiffness Regularization}
\label{ssec:stiffness}
We replace the spatially uniform regularization weights used in prior INR
registration work with a per-voxel stiffness map derived semi-automatically from the CT volume. Each voxel receives a \emph{continuous} stiffness scalar $\kappa$ that follows its CT Hounsfield value and spans three orders of magnitude, from $\kappa\!\approx\!0.01$ in air, through $\kappa\!\approx\!1$ in soft tissue, to $\kappa\!\approx\!100$ in bone, with morphological smoothing and hole-filling so that stiffness varies gradually across tissue interfaces rather than jumping at hard thresholds.
This scalar $\kappa_i$ acts as a continuous material mask that gates how rigidly each point moves with its neighborhood. Rather than penalizing displacement magnitude, we penalize each point's deviation from the $\kappa$-weighted \emph{rigid} consensus motion $\mathbf{T}^\star\!\in\!SE(3)$ of the stiff structure, in proportion to $\kappa_i$:
\begin{equation}
  \mathcal{L}_{\text{mat}} = \frac{\sum_i \kappa_i \bigl\lVert \mathbf{d}(\mathbf{x}_i)-\mathbf{T}^\star(\mathbf{x}_i)\bigr\rVert^2}{\sum_i \kappa_i},
  \qquad
  \mathbf{T}^\star=\operatorname*{arg\,min}_{\mathbf{T}\in SE(3)}\sum_i \kappa_i \bigl\lVert \mathbf{d}(\mathbf{x}_i)-\mathbf{T}(\mathbf{x}_i)\bigr\rVert^2 
  \label{eq:mat}
\end{equation}
where $\mathbf{T}^\star$ is the $\kappa$-weighted least-squares rigid fit to the current field, solved in closed form each iteration. A bone point ($\kappa=100$) is thus driven onto the rigid consensus---zero shape change, while remaining free to translate and rotate---whereas soft tissue ($\kappa\approx1$) is left essentially unconstrained and air ($\kappa\approx0$) drops out. A Jacobian regularizer $\mathcal{L}_{\mathrm{reg}}=\frac{1}{N}\sum_i\bigl[\lambda_F\lVert\mathbf{J}_i\rVert_F^2 + \lambda_d\,w(\kappa_i)\,(\det(\mathbf{I}+\mathbf{J}_i)-1)^2\bigr]$ is retained for global smoothness and fold prevention, where the folding (determinant) penalty is itself stiffness-weighted by $w(\kappa_i)=1/(100.1-\kappa_i)$ so that volume change is suppressed most strongly inside stiff structures.

\subsection{Contact-Zone Displacement Prior}
\label{ssec:air}

We introduce a novel loss term that encodes the physical reality of probe
contact: when the probe is pressed against tissue, the crescent-shaped region between its flat face and the concave fan arc must be air. The extended domain $\Gamma$ covers every pixel column between the upper fan horns, from the image top down to the fan's concave arc.

\paragraph{Displacement prior loss.}
The INR is queried at crescent coordinates $\mathbf{x}\!\in\!\Gamma$ each
iteration, and the CT is sampled at the displaced positions:
\begin{equation}
  \mathcal{L}_{\mathrm{con}}
  = \frac{1}{|\Gamma|}\sum_{\mathbf{x}\in\Gamma}
  \Bigl[\mathrm{ReLU}\!\bigl(I_{\mathrm{CT}}(\mathbf{x}+\mathbf{d}(\mathbf{x}))-a_0\bigr)/s\Bigr]^2,
  \label{eq:disp}
\end{equation}
where $a_0$ and $s$ are the air-intensity reference and scale of the normalized CT; the one-sided rectifier penalizes only above-air (tissue-level) intensity remaining in the crescent. Minimizing $\mathcal{L}_{\mathrm{con}}$ drives the field to push CT
coordinates toward low-intensity regions, enforcing downward tissue
compression without any ground-truth annotation.

\subsection{Fan-Geometry-Aware Optimization}
\label{ssec:fan}

We tailor the optimization strategy to the fan-shaped, partial field of view of a convex transducer through three geometric mechanisms: fan-constrained patch sampling, beam-aligned directional regularization, and a smooth boundary weight.

\paragraph{Fan-constrained patch sampling.}
Patch centers are drawn uniformly from within $\Omega_{\mathrm{fan}}$,
ensuring every queried location has valid spatial neighbors for the NGF gradient computation and no network capacity is spent on background. 

\paragraph{Beam-aligned directional regularization.}
For each pixel inside the fan, a unit beam vector
$\hat{\mathbf{b}}(\mathbf{x})$ is computed from the predefined
fan apex to $\mathbf{x}$, and the in-plane displacement is decomposed into
radial (along-beam) and tangential (cross-beam) parts. An asymmetric penalty
\begin{equation}
  \mathcal{L}_{\mathrm{fan}}
  = \frac{\lambda_{\mathrm{tan}}}{N}\sum_{i}\bigl\|\mathbf{d}_{\mathrm{tan}}^i\bigr\|^2
  + \frac{1}{N}\sum_{i}
  \Bigl(\lambda_r\,\mathrm{ReLU}(d_{\mathrm{rad}}^i)^2
  +\lambda_r^{-}\,\mathrm{ReLU}(-d_{\mathrm{rad}}^i-\tau)^2\Bigr)
  \label{eq:fan}
\end{equation}
encodes the physical prior that the probe acts as a compressor:
retrograde motion (toward the apex) is penalized more strongly than
forward compression, which is allowed freely; a small deadzone $\tau$ leaves minor toward-probe motion unpenalized. 

\paragraph{Smooth boundary weight.}
Predicted displacements are multiplied element-wise by a precomputed map
$m(\mathbf{x})\!\in\![0,1]$, constructed by dilating the fan mask outward,
applying a Gaussian ramp, and multiplying by an angular taper $\cos^2(\theta/\theta_{\max})$ that attenuates displacement toward the lateral fan horns where beam quality is weakest. 

\subsection{Implementation Details}
\label{ssec:impl}
To enhance structural features for the similarity metric, both modalities undergo intensity normalization, and the ultrasound frames are preprocessed using the self-supervised Speckle2Self~\cite{li2025speckle} denoising framework to mitigate signal noise and acoustic artifacts. The SIREN backbone uses 2 layers and 128 hidden units. Raw voxel indices are min-max normalized to $[0,1]^3$ before network input; outputs are rescaled to voxel units before warping via trilinear interpolation~\cite{sitzmann2020implicit,wolterink2022implicit}. The full objective is
$\mathcal{L} = \lambda_{\mathrm{NGF}}\mathcal{L}_{\mathrm{NGF}}
+ \mathcal{L}_{\mathrm{mat}}
+ \mathcal{L}_{\mathrm{fan}}
+ \mathcal{L}_{\mathrm{con}}
+ \mathcal{L}_{\mathrm{reg}}
+ \mathcal{L}_{\mathrm{band}}$,
where $\mathcal{L}_{\mathrm{reg}}$ is the Jacobian regularizer above and $\mathcal{L}_{\mathrm{band}}$ applies the same smoothness penalty to a dilated band just outside the fan, keeping the out-of-view extrapolation continuous and preventing off-plane rigid structures from deforming.
It is optimized with Adam (lr~$3\!\times\!10^{-5}$, gradient clip 1.0) for up to 200 iterations. Validation NGF is evaluated every 50 iterations; optimization stops early if no improvement exceeds $10^{-4}$ for 50 checks, restoring best-seen weights.

\section{Experiments and Results}
\label{sec:experiments}

We evaluate on four robotic US--CT sweeps spanning two acquisition foci: \textbf{Dataset~1}, acquired on a CIRS Model 057A Triple Modality Abdominal Phantom, and \textbf{Dataset~2}, acquired on a Kyoto Kagaku Dual Modality Human Abdomen Phantom (Model US-22); both were acquired with a robot-assisted setup~\cite{li2025robotic}. Manual landmarks were annotated on five evenly spaced frames per sweep ($478$ pairs total), and the deformation is run per-frame on the full cine. Ultrasound (US) data were captured at $880 \times 660$ resolution using a Siemens ACUSON Juniper (5C1 probe). For CT, a mobile "ImagingRing" (medPhoton, Austria) was used. The robotic systems were co-registered via hand-eye calibration using the Ring's integrated optical tracker.


\paragraph{Metrics.}
Registration accuracy is measured with the \emph{(i)} Target Registration Error~(TRE), the mean Euclidean distance between corresponding manually placed landmarks after registration ($478$ landmark pairs across all sweeps; significance vs.\ rigid by Wilcoxon signed-rank test). Image agreement is measured by \emph{(ii)} phase-congruency NCC~(PC)~\cite{kovesi1999}, a contrast-invariant US--CT similarity measure (higher~=~better); we deliberately use a phase-based rather than a gradient-based similarity metric here, as the latter mirrors our NGF data term and would bias the evaluation in our favor. Plausibility is measured by \emph{(iii)} the Jacobian folding fraction, the percentage of pixels with $\det(\mathbf{I}+\mathbf{J})<0$ (lower~=~better, zero is ideal), and---on the bone-adjacent sweeps---\emph{(iv)} the bone/soft strain ratio, the mean shape-change $\lVert\nabla\mathbf{d}\rVert$ inside bone relative to soft tissue (lower~=~bone held more rigid; $1$~=~bone deforms like soft). As a secondary appearance measure we additionally report the Dice overlap of a set of corresponding anatomical structures delineated in both modalities.

\paragraph{Baselines.}
We compare against \emph{(i)}~rigid initialization only (no deformable stage); \emph{(ii)}~Demons~\cite{thirion1998image}, a classical diffeomorphic method; \emph{(iii)}~B-spline FFD~\cite{klein2010elastix} with two data terms, local cross-correlation (FFD-LCC) and normalized gradient fields (FFD-NGF); and \emph{(iv)}~Farneb\"ack optical flow. To justify our NGF data term, we also report our own network driven instead by \emph{(v)}~MIND-only~\cite{heinrich2012mind} and \emph{(vi)}~LC2-only~\cite{fuerst2014}. All methods share identical rigid pre-alignment and run per-frame.

\subsection{Quantitative Results}\label{sec:results}

\begin{figure}[t]
  \includegraphics[width=\textwidth]{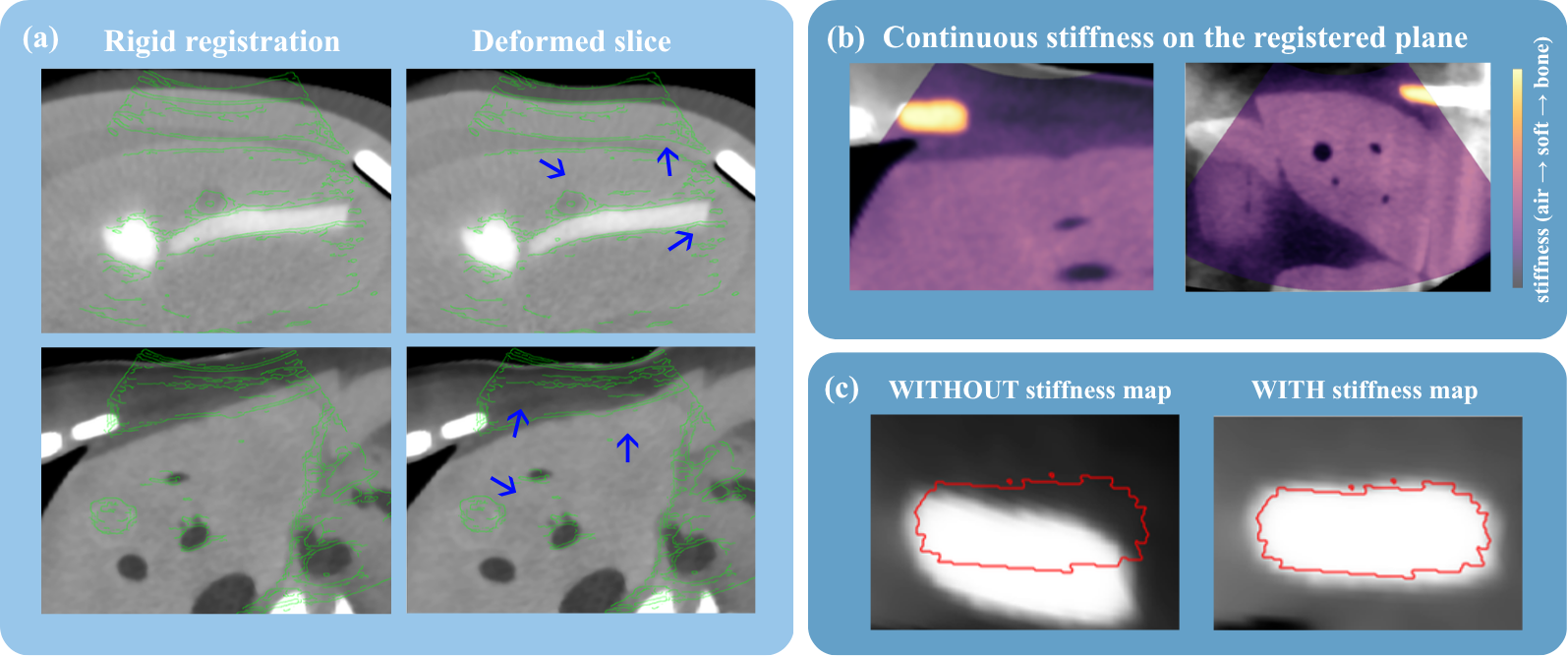}
  \caption{ (a) Qualitative assessment of the stiffness-constrained deformable registration. (b) By incorporating a stiffness map context, the model corrects for transducer-induced deformation, providing the sub-millimeter structural alignment necessary for guided interventions and planning. (c) Note the stability of the bone relative to the compliant soft tissue. }
  \label{fig:results}
\end{figure}

Tab.~\ref{tab:results} summarizes performance pooled across all four sweeps (478 landmarks). Our method attains the lowest TRE ($3.97$~mm; $p<10^{-10}$ vs.\ rigid), outperforming the strongest baseline (FFD-NGF, $4.43$~mm) and all others, while producing essentially fold-free deformations ($0.03\%$ vs.\ $1.4$--$5.6\%$ for FFD/Demons/optical-flow). FFD-NGF attains a marginally higher phase-congruency similarity ($0.053$ vs.\ our $0.048$) and a higher Dice by deforming more aggressively---but at the cost of pervasive folding ($1.41\%$) and with no notion of tissue material, i.e.\ greater appearance similarity without better landmark accuracy. Replacing our NGF data term with MIND or LC2 loses all significance against rigid ($p=0.08$ and $0.88$), confirming NGF as the appropriate US--CT driver. Fig.~\ref{fig:results} shows qualitative results on soft-tissue and bone-adjacent frames.

\begin{table}[t]
  \centering
  \caption{Quantitative comparison, pooled over all four sweeps ($478$ landmarks). The $p$ column reports a Wilcoxon signed-rank test vs.\ rigid; all deformable methods improve significantly ($p<0.05$) except MIND-/LC2-only (n.s.), while Farneb\"ack is significantly \emph{worse}. PC: phase-congruency NCC. Best in each column in \textbf{bold}, second best \underline{underlined}.}
  \label{tab:results}
  \scalebox{0.85}{
    \begin{tabular}{lccccc}
      \toprule
      \textbf{Method} & PC $\uparrow$ & Dice $\uparrow$ & Fold\% $\downarrow$ & $p$ vs.\ rigid & TRE [mm] $\downarrow$ \\
      \midrule
      Rigid only                           & --    & --    & ---   & -- & 4.79 \\
      Demons \cite{thirion1998image}       & 0.045 & \underline{0.56}  & 2.39 & $1\!\times\!10^{-7}$ & 4.59 \\
      FFD-LCC \cite{klein2010elastix}      & 0.032 & 0.54  & 0.29 & $0.02$ & 4.50 \\
      FFD-NGF \cite{klein2010elastix}      & \textbf{0.053} & \textbf{0.65} & 1.41 & $7\!\times\!10^{-5}$ & \underline{4.43} \\
      Farneb\"ack                          & 0.012 & 0.45  & 5.58 & $4\!\times\!10^{-22}$ & 5.85 \\
      MIND-only \cite{heinrich2012mind}    & 0.037 & 0.48 & \underline{0.03} & $0.08$ & 4.63 \\
      LC2-only \cite{fuerst2014}           & 0.025 & 0.48  & \textbf{0.02} & $0.88$ & 4.80 \\
      \midrule
      \textbf{Ours}                        & \underline{0.048} & 0.55 & \underline{0.03} & $6\!\times\!10^{-14}$ & \textbf{3.97} \\
      \bottomrule
  \end{tabular}}
\end{table}

\subsection{Ablation Study}
\label{ssec:ablation}

Tab.~\ref{tab:ablation} isolates each component. Removing the material term $\mathcal{L}_{\mathrm{mat}}$ leaves TRE essentially unchanged ($3.95$ vs.\ $3.97$~mm) yet the bone/soft strain ratio jumps from $0.58$ to $1.10$: the material term enforces bone rigidity \emph{at no accuracy cost}, which is precisely its intended role---soft tissue deforms while the rib is held rigid. Removing the contact/displacement prior $\mathcal{L}_{\mathrm{con}}$ degrades TRE by $0.21$~mm and lowers phase-congruency similarity. $\mathcal{L}_{\mathrm{fan}}$ is TRE-neutral, as it constrains only the near-field region that contains no landmarks. Finally, substituting the NGF data term with MIND or LC2 collapses accuracy to no-better-than-rigid ($4.63$/$4.80$~mm, n.s.), the strongest evidence for the NGF choice. Landmarks were placed by a clinically trained annotator ($\sim$20--29 points per frame).

  

\begin{table}
  \centering
  \caption{Ablation of the loss terms and the data term (pooled over all sweeps; bone/soft strain on the bone-adjacent sweeps). PC: phase-congruency NCC; $p$: Wilcoxon vs.\ rigid. The material term $\mathcal{L}_{\mathrm{mat}}$ drives the bone/soft ratio well below $1$ at no TRE cost; removing it returns the ratio to $\approx\!1$ (bone deforms like soft). Best in each column in \textbf{bold}, second best \underline{underlined}.}
  \label{tab:ablation}
  \small
  \begin{tabular}{l ccccc}
    \toprule
    \textbf{Configuration} & PC $\uparrow$ & bone/soft $\downarrow$ & Fold\% $\downarrow$ & $p$ vs.\ rigid & TRE [mm] $\downarrow$ \\
    \midrule
    w/o $\mathcal{L}_{\mathrm{mat}}$ & \textbf{0.049} & 1.10 & 0.03 & $6\!\times\!10^{-15}$ & \textbf{3.95} \\
    w/o $\mathcal{L}_{\mathrm{con}}$ & 0.036 & \textbf{0.43} & \textbf{0.00} & $7\!\times\!10^{-13}$ & 4.18 \\
    w/o $\mathcal{L}_{\mathrm{fan}}$ & \underline{0.048} & 0.60 & 0.03 & $5\!\times\!10^{-13}$ & \underline{3.96} \\
    MIND data term & 0.037 & --   & 0.03 & $0.08$ & 4.63 \\
    LC2 data term  & 0.025 & --   & \underline{0.02} & $0.88$ & 4.80 \\
    \addlinespace
    Full model (ours) & \underline{0.048} & \underline{0.58} & 0.03 & $6\!\times\!10^{-14}$ & 3.97 \\
    \bottomrule
  \end{tabular}
\end{table}

\section{Conclusion}
\label{sec:conclusion}

We presented a deformable CT-US registration framework based on implicit neural representations of the displacement field, augmented with three anatomy-aware constraints. Together these components address the two principal failure modes of existing INR registration methods in the CT–US setting, which are non-physical deformation of rigid structures and unconstrained behavior outside the imaging field of view. Our results demonstrate improved alignment over rigid initialization and classical deformable baselines on both soft-tissue and bone-containing phantoms, with near-zero topological folding throughout. Several limitations remain. 
The stiffness map relies on a CT-intensity-to-stiffness mapping calibrated for the phantom that may generalize less well to pathological tissue or implants, and the fan-geometry assumptions are specific to convex transducers. A significant challenge in US-CT registration is the limited reliability of similarity metrics, which often fail to capture complex morphological correspondences. To address this, we relied on expert-validated anatomical landmarks to calculate the TRE. While our method demonstrates high qualitative and quantitative alignment with phantom structures, we acknowledge that the transition to \emph{in vivo} studies may introduce greater anatomical variability. A robust, modality-independent registration metric is needed to reduce reliance on manual landmark selection and mitigate potential inter-observer bias, and while probe placement makes simultaneous CT-US capture challenging, future work would benefit from ground truth volumetric deformations based on volumetric imaging for comprehensive validation.

\bibliographystyle{splncs04}
\bibliography{ref}

\end{document}